# The System Description of dun_oscar team for The ICPR MSR Challenge

Binbin Du*, Rui Deng*, Yingxin Zhang*

*Yidun AI Lab, Netease Hangzhou 310052*

{dubinbin, zhangyingxin03, dengrui01 }@corp.netease.com

**Abstract**

*This paper introduces the system submitted by dun_oscar team for the ICPR MSR Challenge. Three subsystems for task1-task3 are descripted respectively. In task1, we develop a visual system which includes a OCR model, a text tracker, and a NLP classifier for distinguishing subtitles and non-subtitles. In task2, we employ an ASR system which includes an AM with 18 layers and a 4-gram LM. Semi-supervised learning on unlabeled data is also vital. In task3, we employ the ASR system to improve the visual system, some false subtitles can be corrected by a fusion module.*

## 1. Introduction

Extracting subtitles in videos is a challenging task, due to the clutter background texts in visual modality and the low speech quality in audio modality. Though the task is difficult, it is not a popular research problem. Many previous studies are only focused on OCR、ASR, which is a core part of the system of extracting subtitles.

In OCR domain, many algorithms are proposed to solve the difficult problems in detection and recognition. EAST[1], PSENet[2], DBNet[3], SAST[4] are both recent popular OCR detection algorithms, they are widely used for the computation efficiency or the ability to detect texts with arbitrary shape. We employ SAST[4] algorithm in our detection model. Except for the usual text center line segmentation branch, SAST utilizes other three segmentation branches to localize texts more precisely.

Two class common methods for OCR recognition are: 1) CTC-based, such as CRNN[5], STAR-Net[6], EnEsCTC[7]; 2) Attention-based, such as ASTER[8], MASTER[9]. Attention-based method usually achieves better results in public research datasets, due to its ability to handle text with irregular shape. But in the video subtitle scenario, the shape of text is usually regular rotated box. Due to the good performance of CTC-based method for Chinese characters, we employ CTC-based method to recognize text content in visual modality.

In ASR domain, Hybrid Attention/CTC training framework[10] is most popular among End-to-End(E2E) automatic speech recognition systems, we also employ the training framework in our system.

Methods in ASR inference stage diversify. [10] utilizes beam search on attention-based branch, and rescore the n-best results with CTC branch. Wenet[11] adopts a different decoding method, beam search are applied in CTC branch firstly, and rescore the n-best results with attention branch. In all methods, a powful LM is very helpful. In our system, we employ a decoding method like[11]. A 4-gram LM is used to rescore in the decoding stage.

In task1, extracting subtitles in visual modality, a OCR model is necessary but enough, we need a module to extract the real subtitles, and discard the cluttered background texts. In this task, we develop a text tracker and a NLP classification model to achieve this goal. The training corpus of NLP model comes from the filtered text by the text tracker. The text tracker can identify most non-subtitles, and the NLP model can strengthen the ability.

For task2, extracting subtitles in audio modality, an ASR system can satisfies the requirement. To achieve better results, we design the inference method as mentioned before. In training stage, strong data augmentation and semi supervised learning[12] are conducted.

For task3, extracting subtitles with both visual and audio modality, we employ a fusion module to integrate two results that in single modality. Subtitles in visual modality are extracted firstly, then subtitles in audio modality are extracted, at last the audio subtitles are used to improve visual subtitles by removing non-subtitles, inserting missing subtitles, and correcting false recognized subtitles.

---

* Equal contributions to this work

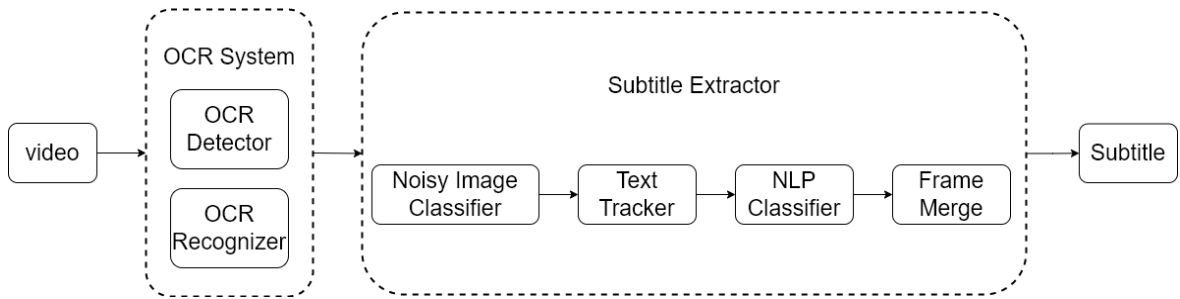

Figure 1. Extracting subtitles in visual modality

## 2. Extracting subtitles in visual modality

The system for task1 consists of a OCR module and a subtitle extractor module. The pipeline are shown as Fig 1.

### 2.1. OCR module

We choose SAST algorithm[4] with a Resnet50 backbone to build our detector. As mentioned in[4], various geometric properties of text regions, including text center line, text border offset, text center offset, and text vertex offset are adopted to ease training and reconstruct the precise polygon of a text instance.

For recognizer, we choose CTC-based method. We employ ResnetXT-18 as the backbone network, and two multi head self-attention layer are inserted before classification layer.

### 2.2. Subtitle extractor module

After OCR module, the all texts in video will be detected and recognized. A subtitle extractor to distinguish subtitles and non-subtitles is necessary.

To achieve the goal, we employ a simple image classifier to filter some simple non-subtitles at first. The classifier accepts single image as input, and outputs the probability of whether the image is a subtitle. And a high frequency rule also works in this step.

The remained non-subtitle texts are difficult to identify by the single image information. We develop a tracker using Hungarian algorithm to build the connection between texts with similar positions. After the tracker processes whole video, all texts will be divided into several instances located in similar positions.

At last of this module, a NLP classifier with three convolution layers and a FC layer will do the final filtering. After filtering all non-subtitles, the subtitle in each frame will be merged according to their text similarity.

### 2.3. Training data mining

As the requirements of the ICPR MSR competition, only 10K ChineseOCR Synthetic dataset, LSVT dataset[13], the provided training data with audio annotation, and the own synthetic data can be used.

For Synthetic data, we first select 10K data form ChineseOCR dataset, the rule to select data is to maximize the amount of different characters. When synthesizing our own data, we use the provided corpus to obtain 100k data with simplified Chinese characters, 100k data with traditional Chinese characters, and 20k data with special punctuations.

In LSVT dataset, there are 30k training data with full annotations and 400k training data with weak annotations. The weak annotations only consist of transcribed texts, but not the location of texts. The data with full annotations can be used to train our OCR detection and recognition model directly. The data with weak annotations can be assigned with strong annotations by comparing with the recognized results generated by our OCR model. By iteratively updating our model and mining weak labeled data, les than 300k data with weak annotations are assigned to strong label to train our model.

The mining procedure of provided training data with audio annotation in task1 is similar to the mining procedure of LSVT. An initial model can be obtained by the available public data, then the mining procedure will be done iteratively.

## 3. Extracting subtitles in audio modality

The system for task2 is a pure ASR system.

### 3.1. Acoustic model

Acoustic model is the most important part in an E2E ASR system. Conformer[14] is a popular model architecture and achieves competitive results on many public datasets. And some researches depict that the bigger the model, the smaller CER[15]. But for this challenge, the amount of training data is limited, big model may overfits the train set, and achieves worse result in the test stage. To alleviate the problem, stochastic depth[16] is applied in our acoustic model training stage.

Our acoustic model adopts Conformer as our encoder architecture, the attention dim is 512, the number of head is 16, the number of layers is 18.

### 3.2. Decoder

An ASR system with a fine decoder usually performs better compared with a pure acoustic model. In our system, a decoder like that in [11] is employed in our system. Beam search is conducted in CTC branch firstly, and attention branch is applied to rescore the n-best results.

The previous decoding procedure only works on the different branches of the acoustic model, an extra LM can be also employed in this stage. To improve the adaptability to the TV show domain, a universal LM trained by common corpus and a domain specified LM trained by the corpus in this challenge will work simultaneously.

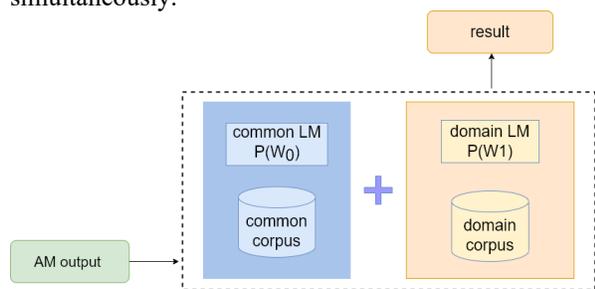

Figure 2. Multi-LM decoder

### 3.3. Training data mining

As the requirements of this competition, only aishell1 training data[17] and the provided training data with visual annotation can be used. The provided taring data need to be converted corresponding audios and transcribed texts.

The provided annotations are framewise, there are many repeated contents in it. Firstly we merge the frame-wise annotations like Fig 3, then we employ an ASR system trained by aishell1 data to filter the non-subtitles. The data mining procedure is conducted with four rounds.

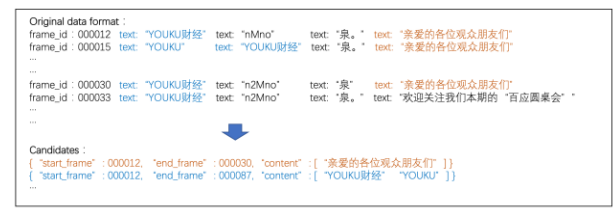

Figure 3. Merge the framewise annotations

Except for the data with various annotations, 200h unlabeled data is provided. Semi supervised learning on this data is also useful to improve accuracy. A method like [12] is conducted with six rounds, as shown in Fig4.

## 4. Extracting subtitles with both visual and audio modality

The system for task3 is most complicated, we employ a subtitle extractor with visual modality, a subtitle extractor with audio modality, and a fusion module to integrate two results in single modality.

The two subtitle extractors in single modality are developed by the method described in previous two chapters. The fusion module is shown in Fig 5. A filter, a remover and a padder works in the module.

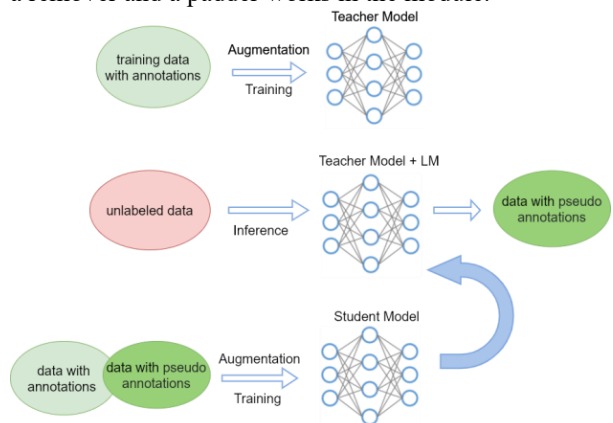

Figure 4. Semi-supervised learning

The fusion module combines two modalities at result-level, it is more flexible than that combining at model-level. The visual subtitle extractor produces some subtitles, which have been filtered by the subtitle tracker and the NLP classifier.

The produced subtitles are processed by a splitter firstly, some subtitles that be mistakenly merged in visual system can be split with reference to the Asr results, like Fig6.

Figure 6. Splitter in the fusion module

There is a filter in this system to filter false candidates. If one subtitle set cross multi frames has several different candidate contents, we will compare the visual subtitle with the audio subtitle, the similarities of characters and syllables are considered to rank the candidates. The common LM score also is employed. Some examples are shown in Fig 7.

The remover in the fusion module will remove some subtitles with low similarity to ASR results. And the padder will insert some missing subtitles in visual modality, like Fig 8.

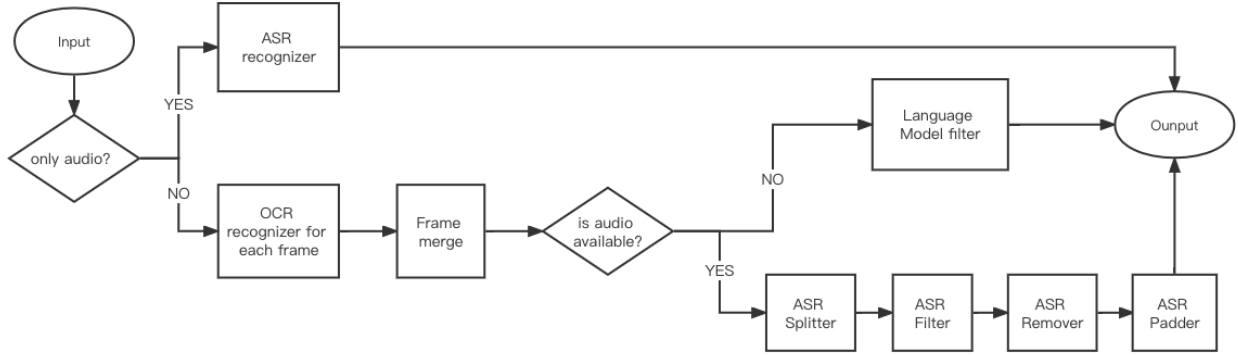

Figure 5. Fusion module that combines two modalities

Figure 7. Filter in the fusion module

Figure 8. Padder in the fusion module

## 5. Experiments and Results
### 5.1. Experimental Setup

The training data of OCR model consists four parts: (1) LSVT dataset with full annotations and weak annotations; (2)10K ChineseOCR dataset; (3) the provided training data; (4) 220K our synthetic data. The training data of ASR model consists two parts: (1) aishell1 training dataset; (2) the provided training data. The provided training data in various tasks is only used in the specified task.

The backbone of OCR detector is Resnet50, and the backbone of OCR recognizer is ResnetXT-18 with two MHSA layers. The threshold of the image classifier in the subtitle extractor is 0.05.

The backbone of ASR acoustic model is Conformer, the attention dim is 512, the number of head is 16, the number of layers is 18. The common LM and the domain LM are both 4-gram.

### 5.2. Results on three tasks

The results of the three tasks are listed in Tab 1.

Table 1. Results of the three tasks

| Task | Dataset | CER |
|---|---|---|
| Task1 | validation | 0.1692 |
|  | test | 0.1159 |
| Task2 | validation | 0.1851 |
|  | test | 0.2162 |
| Task3 | validation | 0.1604 |
|  | test | 0.1319 |

## 6. Conclusions

In the visual system, a OCR model with a SAST detector and a CTC-based recognizer is developed. And a subtitle extractor is conducted to distinguish subtitles and non-subtitles. In the audio system, a well designed ASR system is completed to extract all subtitles in audio modality.

In the most complicated task3, except for the separate audio and visual systems, a fusion module is designed to handle some hard cases in single modality.

Combining the submitted results on the validation set and the test set, we task the third, third and first place in three tasks respectively.